\documentclass[lettersize,journal]{IEEEtran}
\usepackage{amsmath,amsfonts}
\usepackage{algorithmic}
\usepackage{array}
\usepackage[caption=false,font=normalsize,labelfont=sf,textfont=sf]{subfig}
\usepackage{textcomp}
\usepackage{stfloats}
\usepackage{url}
\usepackage{verbatim}
\usepackage{graphicx}
\def\BibTeX{{\rm B\kern-.05em{\sc i\kern-.025em b}\kern-.08em
    T\kern-.1667em\lower.7ex\hbox{E}\kern-.125emX}}
\usepackage{balance}

\usepackage{xspace}
\usepackage{booktabs}
\usepackage[numbers,sort&compress]{natbib}
\newcommand{\eg}{{\emph{e.g.}}}
\newcommand{\ie}{{\emph{i.e.}}}
\usepackage[normalem]{ulem} 
\useunder{\uline}{\ul}{}

\newcommand{\corr}[1]{\textcolor{black}{#1}}

\usepackage[switch]{lineno}

\usepackage{amssymb}

\usepackage{mathbbol}
\usepackage{multirow}
\usepackage{graphicx}
\usepackage{algorithm}
\usepackage{algorithmic}
\usepackage{amssymb}
\usepackage{amsmath}
\usepackage{wrapfig}
\usepackage{caption}
\usepackage{subcaption}
\usepackage{multirow}

\usepackage{microtype}      
\usepackage{xcolor}         
\newcommand{\name}{{Chimera-Seg}}
\usepackage[colorlinks=true, urlcolor=blue]{hyperref}

\begin{document}
\title{Partial CLIP is Enough: \\ \includegraphics[width=0.5cm]{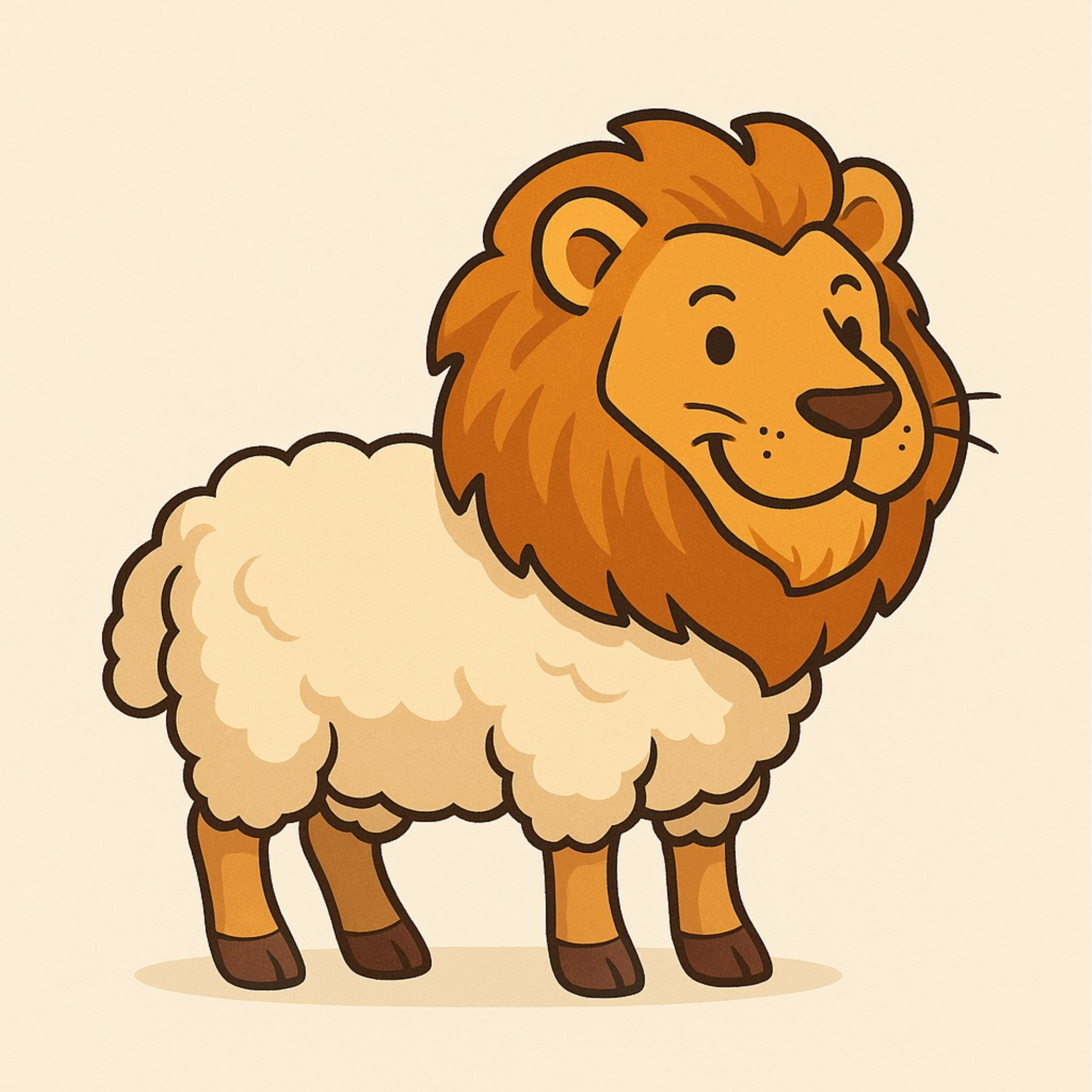} Chimera-Seg for Zero-shot Semantic Segmentation}
\author{
Jialei Chen$^*$, Xu Zheng$^\dagger$, Danda Pani Paudel, Luc Van Gool, Hiroshi Murase,~\IEEEmembership{Life Fellow,~IEEE} Daisuke Deguchi
\thanks{Jialei Chen, Daisuke Deguchi, Hiroshi Murase are with the Graduate School 
of Informatics, Nagoya University, Nagoya, 
Japan. Xu Zheng is with AI Thrust, The Hong Kong University of Science and Technology, Guangzhou Campus (HKUST-GZ), Guangzhou, 
China. 
Danda Pani Paudel, Luc Van Gool are with the INSAIT, Sofia University, St. Kliment Ohridski.
* indicates the corresponding author. $\dagger$ indicates the project leader.}

}

\markboth{Journal of \LaTeX\ Class Files,~Vol.~18, No.~9, September~2020}%
{How to Use the IEEEtran \LaTeX \ Templates}

\maketitle

\begin{abstract}
Zero-shot Semantic Segmentation (ZSS) aims to segment both seen and unseen classes using supervision from only seen classes. Beyond adaptation-based methods, distillation-based approaches transfer vision-language alignment of vision-language model, \eg, CLIP, to segmentation models. However, such knowledge transfer remains challenging due to: \textbf{(1)} the difficulty of aligning vision-based features with the textual space, which requires combining spatial precision with vision-language alignment; and \textbf{(2)} the semantic gap between CLIP’s global representations and the local, fine-grained features of segmentation models. To address challenge \textbf{(1)}, we propose \name, which integrates a segmentation backbone as the body and a CLIP-based semantic head as the head, like the Chimera in Greek mythology, combining spatial precision with vision-language alignment. Specifically, \name\ comprises a trainable segmentation model and a \textit{CLIP Semantic Head} (CSH), which maps dense features into the CLIP-aligned space. The CSH incorporates a frozen subnetwork and fixed projection layers from the CLIP visual encoder, along with lightweight trainable components. \corr{The partial module from CLIP visual encoder, paired with the segmentation model, retains segmentation capability while easing the mapping to CLIP’s semantic space.} To address challenge \textbf{(2)}, we propose \textit{Selective Global Distillation (SGD)}, which distills knowledge from dense features exhibiting high similarity to the CLIP CLS token, while gradually reducing the number of features used for alignment as training progresses. Besides, we also use a \textit{Semantic Alignment Module (SAM)} to further align dense visual features with semantic embeddings extracted from the frozen CLIP text encoder. Experiments on two benchmarks show improvements of 0.9\% and 1.2\% in hIoU.
\end{abstract}


\section{Introduction}
As one of the most significant tasks in computer vision, semantic segmentation \cite{FCN,segmentationTCSVT,frozen,zheng2023both,zheng2024learning,zheng2025distilling} has achieved great advancement with the development of deep learning \cite{resnet, transformer}. However, achieving high performance segmentation model needs large scale datasets which are very time-consuming and expensive to achieve. Therefore, inspired by zero-shot classification, zero-shot semantic segmentation also achieves great success \cite{sign,zs3,cagnet,spnet}. In the era of rapidly advancing foundation models \cite{clip,sam}, zero-shot learning has gained renewed attention.

\begin{figure}[!t]
\centering

\begin{minipage}{\linewidth}
    \centering
    \includegraphics[width=0.9\linewidth]{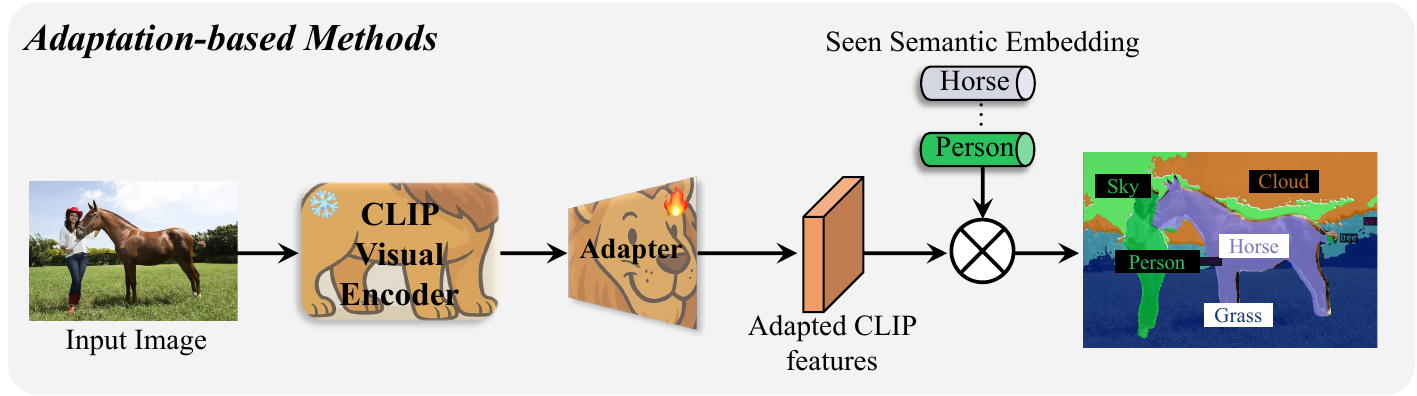}
    \parbox{0.95\linewidth}{
        \small ~~~\textbf{(a)}~Adaptation-based methods.
    }
    \label{fig:existing_methods_a}
\end{minipage}

\vspace{1em}

\begin{minipage}{\linewidth}
    \centering
    \includegraphics[width=0.9\linewidth]{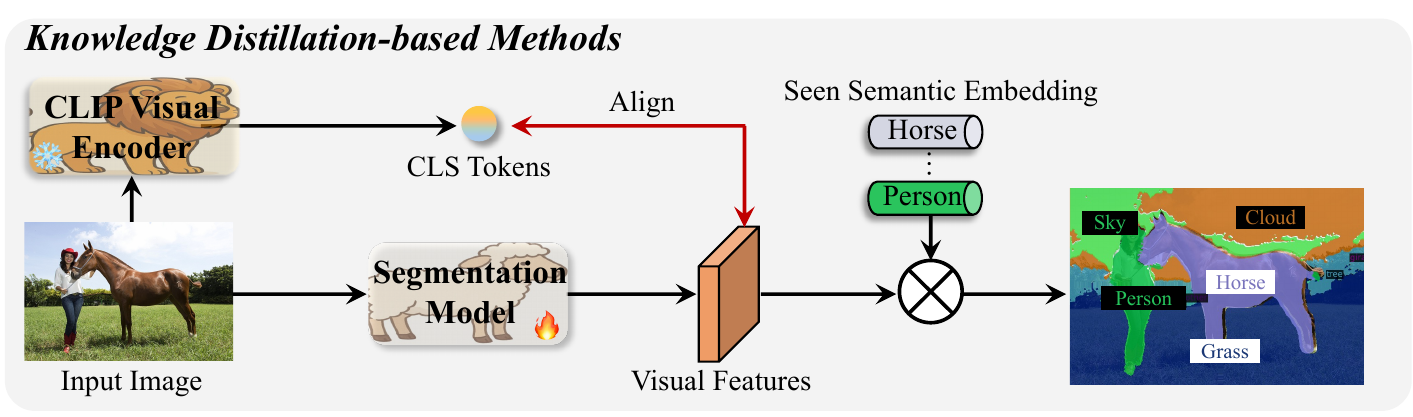}
    \parbox{0.95\linewidth}{
        \small ~~~\textbf{(b)}~Knowledge distillation based methods.
    }
    \label{fig:existing_methods_k}
\end{minipage}

\vspace{1em}

\begin{minipage}{\linewidth}
    \centering
    \includegraphics[width=0.9\linewidth]{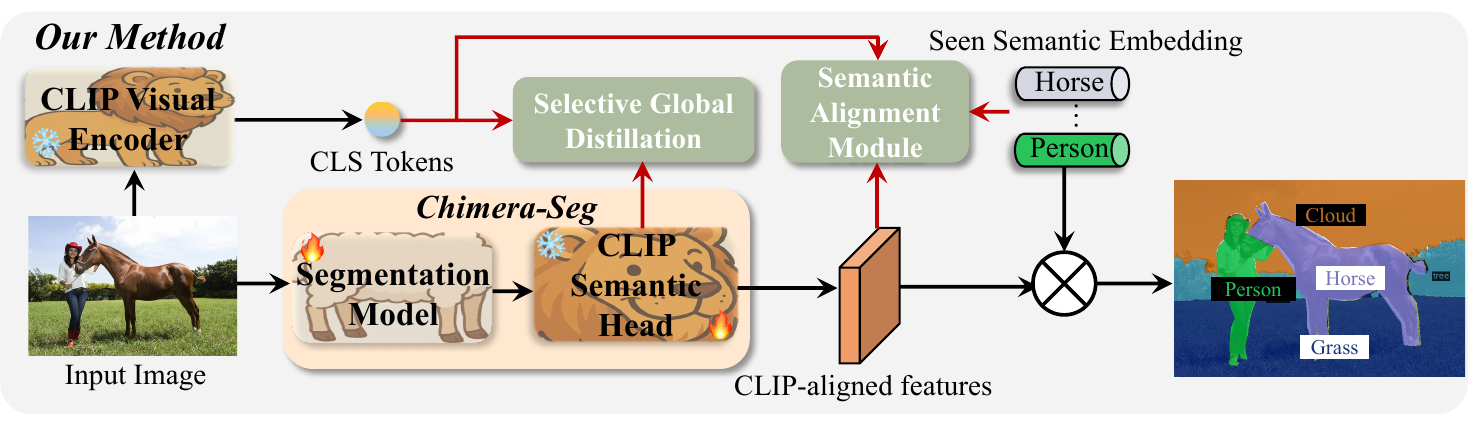}
    \parbox{0.95\linewidth}{
    \vspace{-0.1in}
        \small ~~~\textbf{(c)}~Our method.
    }
    \vspace{-0.1in}
    \label{fig:our_method}
\end{minipage}
\vspace{0.5em}
\caption{Existing methods can be broadly categorized into (a) adaptation-based methods and (b) knowledge-distillation-based methods. However, both methods face their own challenges: (a) is fixed with the CLIP visual encoder. (b) has to solve the segmentation and vision-to-text projection at the same time. Our methods combines the merits from both mehtods.}
\label{fig:teaser}
\vspace{-0.2in}
\end{figure}

A growing body of work focuses on transferring the zero-shot and open-vocabulary capabilities of foundation models, most notably CLIP \cite{clip}, to tasks requiring fine-grained understanding. Existing methods can be boardly categoriezd into two types: adaptation-based and knowledge distillation-based. While early approaches either fine-tuned CLIP \cite{Lseg,Openseg} or enhanced it with prompts and adapters \cite{zegclip,convdies,globalknowledgecalibration,ODISE,catseg}, and later efforts introduced spatial priors to compensate for the limitations of self-attention in fine-grained tasks \cite{maskclip,sclip,DeOP,maskawareclip,simplebaseline,mask2former,maskformer} as shown in Fig. \ref{fig:teaser}(a). Recent work has increasingly adopted knowledge distillation \cite{cliptoseg,generalizableembedding,splitmatching} as a more flexible alternative to tuning-based methods as shown in Fig. \ref{fig:teaser}(b). These approaches transfer CLIP’s vision-language alignment into task-specific segmentation models, which can then operate independently during inference. By decoupling from CLIP’s architecture, distillation enables lightweight and flexible model designs while preserving both global semantics and spatial precision through feature-level alignment.

Though promising, existing knowledge distillation methods still face two fundamental limitations. \textbf{(1)} They struggle to effectively map visual features into the semantic space. Specifically, current approaches often require joint optimization of fine-grained semantic segmentation and vision-language alignment. However, these two tasks rely on inherently different inductive biases: the former prioritizes spatial precision at the pixel level, while the latter emphasizes abstraction of high-level semantics. This conflict complicates optimization and often leads to suboptimal performance in one or both tasks. This challenge calls for a method that retains the spatial precision of segmentation models while effectively leveraging the semantic knowledge of vision-language models, without being hindered by their conflicting objectives. Inspired by the Chimera in Greek mythology, combining the body of a goat for agility and the head of a lion for strength, we propose \name, a hybrid framework that integrates a powerful segmentation model as Chimera's body and a CLIP Semantic Head (CSH) as the head to infuse vision-language alignment from CLIP. Specifically, the CSH takes dense visual features from the segmentation model as input, first linearly projects them, then passes them through a frozen transformer block from the CLIP visual encoder to inherit vision-language alignment. The output is subsequently refined via a trainable batch normalization layer and mapped into the CLIP-aligned space using the original, fixed CLIP MLP projection head.
\corr{By combining the segmentation model with only a partial CLIP visual encoder, the approach preserves the strengths of the backbone while seamlessly incorporating CLIP’s semantic knowledge, enabling efficient cross-modal alignment and improving zero-shot segmentation performance.}

Besides the challenge above, existing methods still face \textbf{(2)} a semantic mismatch between CLIP’s global representations and the local, fine-grained features produced by the segmentation model. Specifically, CLIP is pre-trained for image-level classification and thus excels at extracting global semantic representations of prominent objects. In contrast, semantic segmentation necessitates a more granular understanding across the entire image, including small-scale and background regions. Consequently, aligning dense visual features with CLIP’s global CLS token introduces a semantic misalignment, limiting the effectiveness of the knowledge transfer. To this challenge, we propose Selective Global Distillation (SGD), which extracts semantically meaningful regions by measuring similarity between dense features from the segmentation model and the CLIP CLS token. We employ gumbel-softmax-based top-$K$ sampling to select high-similarity regions and aggregate them into a global representation. As training progresses, the number of selected features decays, reducing reliance on noisy background regions. This distilled global target encourages the backbone to focus on discriminative semantics while preserving alignment with CLIP. Lastly, to further enhance the performance of \name, we use a Semantic Alignment Module (SAM) that strengthens textual supervision by aligning both seen semantic embeddings and learned visual prototypes with the CLIP text encoder’s CLS token. By enforcing distributional consistency between semantics and CLS, as well as features and CLS, SAM enables the model to capture richer linguistic structure and reinforces vision-language alignment beyond static category-level representations.

Unlike adaptation-based methods that adopt the full CLIP visual encoder as the backbone, our framework leverages only a small, frozen subset of CLIP’s visual encoder. Rather than replicating the entire architecture, we focus solely on transferring CLIP’s vision-language matching capability, enabling more flexible and effective design of zero-shot segmentation models. In contrast to existing knowledge distillation methods that align all dense features with the CLIP CLS token, we propose a \corr{decayed top-$K$ selecting} strategy that selectively distills from features most semantically aligned with the CLS token. This targeted alignment mitigates semantic drift and enhances the quality of transferred knowledge. Furthermore, our method goes beyond simply treating CLIP’s semantic embeddings as static classifiers by actively leveraging the latent knowledge encoded within the text encoder. Through our Semantic Alignment Module (SAM), we align both semantic embeddings and visual prototypes with CLS tokens from the text encoder, enabling richer and more consistent vision-language alignment. Our contributions are listed as follows:


    - We propose \name, a simple yet effective framework that projects dense visual features into the text-aligned visual space to improve zero-shot segmentation.
    
    - We introduce two key components: Selective Global Distillation (SGD) and Semantic Alignment Module (SAM), which leverage the alignemnt between CLIP’s visual and textual encoders.
    
    - Extensive experiments on two zero-shot segmentation benchmarks demonstrate that our approach significantly improves performance, especially in hIoU, and advances the state of open-vocabulary segmentation.

\section{Related Work}
\subsection{Semantic Segmentation}
Semantic segmentation~\cite{zheng2025reducing,zheng2024360sfuda++,xie2023adversarial,zheng2023look,FCN,frozen,deeplabv3,semanticmatter,maskformer,mask2former,segformer,setr,upernet,uncertaintyteacher,zhu2023good,zheng2024semantics,zheng2024transformer,zheng2024centering,zhang2024goodsam,zhu2024customize,zhong2025omnisam,zheng2024magic++,zheng2024learning,liao2025benchmarking,zhao2025unveiling} aims to assign a semantic label to every pixel in an image. Early approaches were based on convolutional neural networks (CNNs) \cite{FCN}, which effectively capture local patterns but struggle to model long-range dependencies due to limited receptive fields. To mitigate this, various modules such as atrous convolutions \cite{deeplabv3}, non-local blocks~\cite{non-local}, and pyramid-based fusions have been proposed to enhance global context. The introduction of Vision Transformers (ViTs)~\cite{vit} has marked a paradigm shift by leveraging global self-attention for long-range interaction, leading to strong performance in segmentation tasks \cite{segformer,swin,setr}. However, transformers often incur high computational costs and lack inductive biases such as spatial locality, which can hinder convergence under limited supervision. To combine the best of both worlds, hybrid models have emerged~\cite{agoodstudent}, integrating convolutional priors with transformer-based reasoning.

Despite architectural advances, most methods rely on full supervision, requiring dense annotations for all classes, an impractical assumption in open-world scenarios. We instead consider zero-shot semantic segmentation (ZSS), where only seen classes are labeled during training, and the goal is to segment both seen and unseen classes during inference. To tackle this, we propose a framework that leverages pretrained vision-language models (VLMs), selectively distilling their cross-modal knowledge to enhance segmentation under partial supervision. Our approach offers a scalable and annotation-efficient solution toward open-vocabulary segmentation.

\subsection{Zero-shot and Open-vocabulary Semantic Segmenatation}
Though impressive, conventional segmentation models require large-scale datasets with high-quality pixel-wise annotations, which are costly and labor-intensive to obtain. To alleviate this burden, the task of Zero-shot Semantic Segmentation (ZSS) has attracted increasing attention, drawing inspiration from zero-shot classification~\cite{zeroshotTCSVT,zeroshotlearning2}. In ZSS~\cite{spnet,cagnet,zegformer,sign,strict,zegclip,DeOP,cliptoseg,generalizableembedding,zs3}, the label set is partitioned into seen and unseen classes. Only annotations of seen classes are available during training, while inference requires the model to recognize and segment both seen and unseen classes, posing a more realistic yet challenging setting in semantic segmentation.

In parallel, the emergence of large-scale vision-language models (VLMs), such as CLIP, has enabled a new paradigm known as Open-Vocabulary Semantic Segmentation (OVSS) \cite{opensurvey,globalknowledgecalibration,semanticsam,catseg,dstdet,convdies}. Unlike ZSS, which focuses on generalization within a single dataset, OVSS aims to generalize across datasets: for example, training on COCO-Stuff \cite{coco} and evaluating on ADE20K~\cite{ade20k} without requiring additional annotations. VLMs bridge the semantic gap between vision and language, making it possible to transfer knowledge across domains using text embeddings as supervision signals. Our proposed method is developed primarily for the ZSS setting, where the label supervision is limited. 

\subsection{Knowledge Distillation}
Knowledge Distillation (KD) aims to transfer knowledge from a pre-trained teacher model to a student model~\cite{knowledgesurvey,distillingknowledge,quan2023mawkdn,exploringfromclipvisionencoder,distillingdetr,logtits-basedTCSVT,clipself,vild,pading}. Existing KD strategies can be broadly divided into two classes: feature distillation~\cite{exploringfromclipvisionencoder,vild,distillingdetr,hierarchicalvisual-languageknowledgedistillation,clipself,globalknowledgecalibration}, which encourages the student to mimic the teacher’s feature representations directly, and relation distillation~\cite{pading,globalknowledgecalibration}, which preserves higher-order structural relationships, such as pairwise distances or similarity patterns among embeddings. For example, PADing~\cite{pading} introduces pseudo visual embeddings and enforces that their relational distances reflect those in the text embedding space, ensuring semantic consistency. While most prior works focus on extracting knowledge solely from the CLIP visual encoder~\cite{cliptoseg,generalizableembedding,vild,clipself,exploringfromclipvisionencoder}, they often overlook the valuable konwledge in the CLIP text encoder. In contrast, our method introduces a unified distillation framework that leverages both the CLIP visual and text encoders as complementary supervision signals. By aligning visual features to the vision-language embedding space via the CLIP text projection head, and jointly distilling structural knowledge from the textual modality, our approach facilitates more semantically coherent predictions, especially for unseen classes. 

\begin{figure*}[t]
\centering
\includegraphics[width=0.9\linewidth]{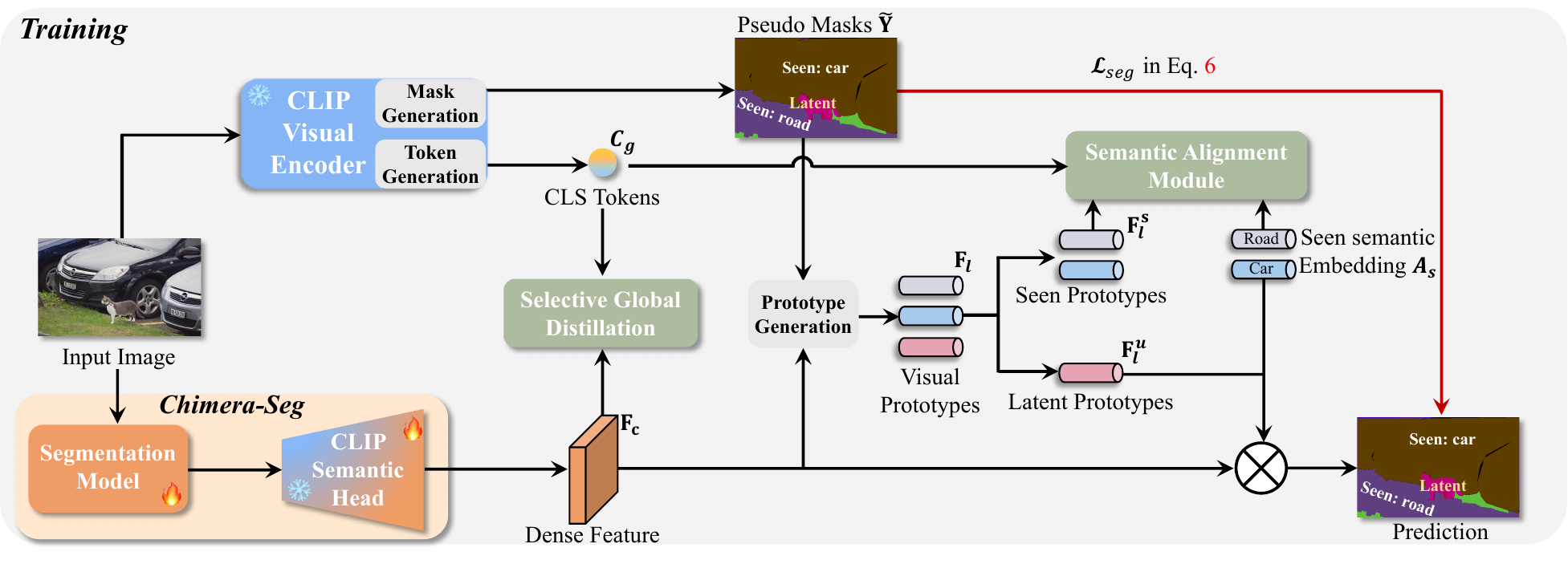}
\vspace{-0.1in}
\caption{Overall framework of our proposed method.
Given an input image, we utilize the CLIP visual encoder to generate pseudo masks $\tilde{\textbf{Y}}$ and CLS tokens $\textbf{C}_g$. The segmentation model extracts dense features, which are aligned with $\textbf{C}_g$ via Selective Global Distillation. The aligned features are then used to generate visual prototypes $\textbf{F}_l$, including seen $\textbf{F}_l^s$ and latent ones $\textbf{F}_l^u$. The semantic alignment module projects both seen and latent prototypes into the semantic space guided by seen-class textual embeddings. The final segmentation prediction is supervised by the pseudo masks, enabling the model to learn transferable representations for both seen and unseen concepts.}
\label{fig:main}
\vspace{-0.2in}
\end{figure*}

\section{Methods}
\subsection{Preliminary and Method Overview}
\noindent \textbf{Preliminary.} Given a dataset $\mathcal{D} = \left\{ \left(\textbf{X}_i,\textbf{Y}_i\right) \right\}_{i=1}^M$, where $\textbf{X}_i \in \mathbb{R}^{H \times W \times 3}$ denotes an input image and $\textbf{Y}_i \in [0,N]^{H \times W}$ its corresponding pixel-level annotations, and a semantic embedding matrix $\textbf{A} \in \mathbb{R}^{N \times D}$ for $N$ classes where $D$ indicates the channel number of semantic embeddings, we first consider the standard closed-set semantic segmentation setting, where both $\textbf{A}$ and $\textbf{Y}_i$ are fully accessible during training. In the zero-shot semantic segmentation (ZSS) setting, the embedding set $\textbf{A}$ is partitioned into seen and unseen subsets, denoted as $\textbf{A}_s \in \mathbb{R}^{N_s \times D}$ and $\textbf{A}_u \in \mathbb{R}^{N_u \times D}$, where $N_s + N_u = N$ and $\textbf{A}_s \cap \textbf{A}_u = \varnothing$. During training, annotations for unseen classes are unavailable, and corresponding pixels are marked as “ignored.” However, since seen and unseen classes often co-occur within the same image, discarding all images containing unseen classes would lead to significant data loss~\cite{spnet,zegformer,zegclip}. Therefore, all images $\textbf{X}_i$ are retained during training. At inference time, ZSS jointly evaluates both seen and unseen classes, in contrast to traditional zero-shot settings that evaluate them separately~\cite{anotherteacher,spnet}. In this work, we adopt the both inductive and transductive ZSS setting.

\noindent \textbf{Method overview.} 
As shown in Fig. \ref{fig:main}, given an input image, we first extract dense features using a trainable backbone (\eg, SegFormer \cite{segformer}). These features are projected into the CLIP-aligned space via CSH, which incorporates a partially frozen CLIP visual encoder. In parallel, the frozen CLIP encoder provides CLS and patch tokens, which are used to generate pseudo masks $\widetilde{\textbf{Y}}$ following~\cite{cliptoseg}, revealing latent class regions. The Selective Global Distillation (SGD) module then refines feature alignment by enhancing similarity between dense features and CLS tokens. Class-wise prototypes $\textbf{F}_l$ are obtained by averaging features within each class region and are split into seen ($\textbf{F}_l^s$) and latent ($\textbf{F}_l^u$) groups. \corr{Following CLIP2Seg \cite{cliptoseg}, latent prototypes are derived from unannotated regions with unknown class categories and counts, and differ from unseen classes, whose names and presence are explicitly defined in the dataset. }The Semantic Alignment Module (SAM) further aligns $\textbf{F}_l^s$ with semantic embeddings $\textbf{A}_s$ from the CLIP text encoder. Final predictions are made by matching prototypes with text embeddings and supervised via pseudo masks.

\begin{figure}[t]
\centering
\includegraphics[width=1\linewidth]{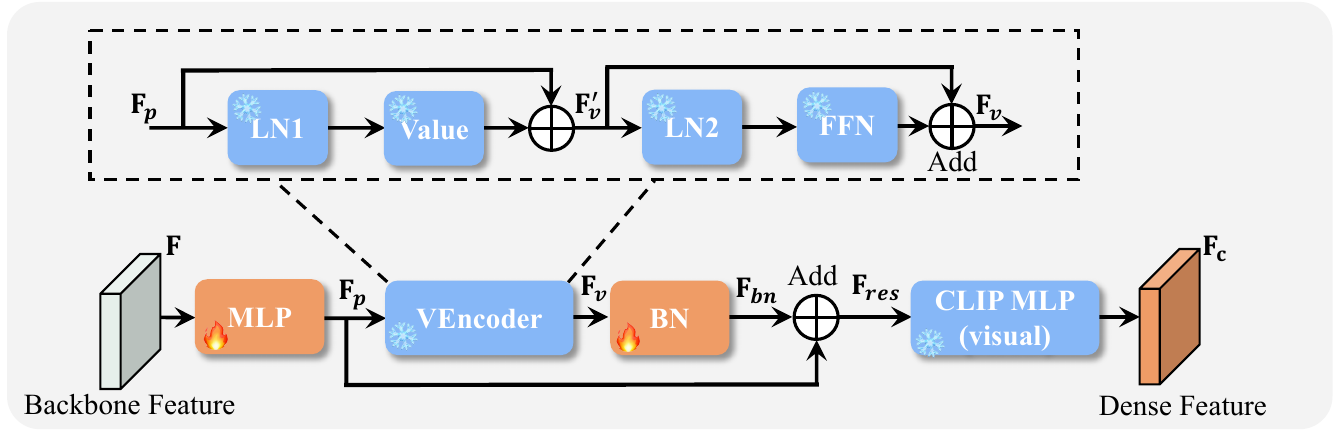}
\vspace{-0.25in}
\caption{The overview of \corr{CLIP Semantic Head (CSH)}.}
\label{fig:vtproj}
\vspace{-0.25in}
\end{figure}

\subsection{\name}\label{sec:VT projector}
Current knowledge distillation methods for ZSS~\cite{cliptoseg,generalizableembedding,splitmatching} aim to transfer vision-language alignment capabilities from CLIP~\cite{clip} to segmentation models such as SegFormer~\cite{segformer}, enabling them to perform semantic segmentation while obtaining the vision-language matching capabilities. However, these methods require the model to jointly learn object-level segmentation and vision-to-language projection, two inherently distinct tasks, which often leads to suboptimal performance. To address this challenge, we propose \textbf{\name}, which leverages fixed components from the CLIP visual encoders to facilitate the projection of visual features into the semantic space. By partially reusing CLIP structures, our approach avoids the computational overhead and spatial interference associated with full CLIP forward passes, while preserving the critical semantic transfer pathway required for effective vision-language alignment. The structure of \corr{\textbf{CLIP Semantic Head (CSH)}} is illustrated in Fig.~\ref{fig:vtproj}. It comprises three main components: (1) a trainable MLP that projects backbone features into the CLIP visual space, (2) a VEncoder that projects input features into the CLIP visual space to facilitate alignment with semantic embeddings, and (3) a frozen MLP taken directly from the final projection layer of the CLIP visual encoder, used to map visual features into semantic-aligned space of CLIP.

Given a feature map $\textbf{F} \in \mathbb{R}^{H \times W \times C}$ extracted from a trainable backbone (\eg, SegFormer~\cite{segformer}) where $C$ indicates the channel number of the backbone features, we first apply a trainable MLP to project the features into CLIP's embedding dimension $D$, resulting in $\textbf{F}_p = \text{MLP}(\textbf{F})$. While this projection matches the dimensionality of CLIP, it does not guarantee semantic or structural compatibility due to the inherent domain gap between ImageNet-pretrained vision encoders and CLIP. To bridge this gap, we further process $\textbf{F}_p$ using a partially frozen Transformer sub-block, referred to as \textit{VEncoder}. This module is constructed by extracting components from the final block of the CLIP visual encoder and includes a first layer normalization (LN1), a value projection layer \corr{(Value)}, a second layer normalization (LN2), and a feed-forward network (FFN). All components are frozen during training to preserve the semantic priors learned by CLIP. Notably, each module in \textit{VEncoder} is directly sourced from the final layer of the CLIP visual encoder. Formally, the forward process proceeds as follows: given the projected feature $\textbf{F}_p$, we first apply LN1 followed by a value projection to obtain the contextualized representation: $\textbf{F}_v' = V(\text{LN1}(\textbf{F}_p)) + \textbf{F}_p$.
The result is further normalized and passed through the FFN with another residual connection: $\textbf{F}_{v} = \text{FFN}(\text{LN2}(\textbf{F}_v')) + \textbf{F}_v'$.
This process retains only the value projection and FFN branches from the original CLIP transformer block. Notably, we discard the query-key attention mechanism and preserve only the value projection, following findings from recent studies~\cite{maskclip,sclip} that CLIP’s global self-attention lacks spatial precision and is suboptimal for dense prediction. In contrast, the value branch retains rich semantic content, making it more suitable for vision-language alignment. Spatial structure is maintained by the segmentation backbone, while VEncoder serves as a lightweight, semantics-preserving projection module.
\begin{figure}[t]
\centering
\includegraphics[width=0.9\linewidth]{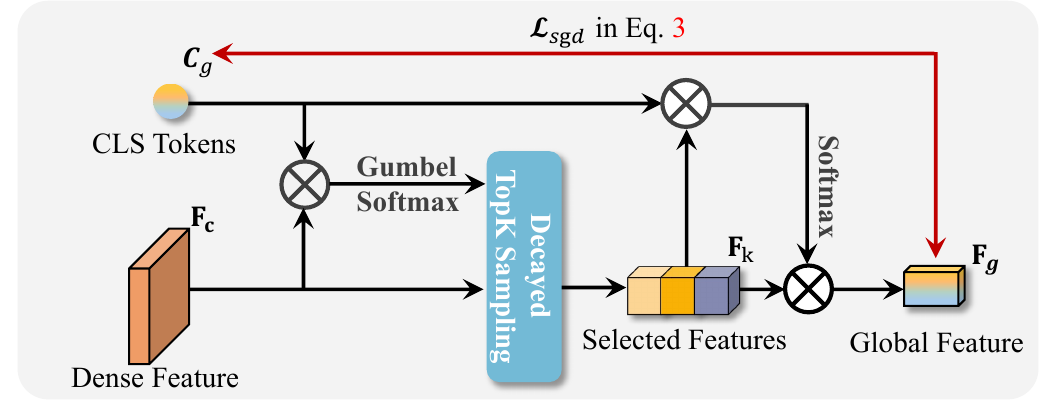}
\vspace{-0.1in}
\caption{Overview of \corr{Selective Global Distillation (SGD)}.}
\label{fig:SGD}
\vspace{-0.25in}
\end{figure}

The output of the VEncoder $\textbf{F}_v$ is subsequently passed through a trainable Batch Normalization layer, yielding $\textbf{F}_{bn} = \text{BN}(\textbf{F}_v)$. This normalization step is introduced to align the feature statistics more closely with those observed in the deeper layers of the CLIP visual encoder, as further demonstrated in Sec.~\ref{sec: ablation}. To improve gradient propagation and stabilize the training process, we incorporate a residual connection by summing the original projection $\textbf{F}_p$ with the normalized output, resulting in $\textbf{F}_{res} = \textbf{F}_p + \textbf{F}_{bn}$. To better align visual features with semantic embeddings that represent class concepts in the dataset, $\textbf{F}_{res}$ is then passed through a frozen MLP head taken from the CLIP visual encoder, which was originally designed for projecting text embeddings into the CLIP semantic space. This step produces the CLIP-bridged feature: $\textbf{F}_{c} = \text{MLP}_{vis}(\textbf{F}_{res})$. By retaining this text-side projection module, we ensure that the visual features are embedded within the same semantic space as CLIP’s semantic embeddings, thereby maintaining intrinsic alignment and enabling seamless interaction with language-driven prototypes.

During training, only the initial MLP and the BN layer are updated, while both the CLIP VEncoder and the visual MLP remain frozen. This design allows the model to benefit from CLIP’s pretrained vision-language alignment while introducing minimal trainable parameters to project the backbone features to the CLIP visual and semantic space, thereby enhancing flexibility and domain adaptability.

\begin{figure}[t]
\centering
\includegraphics[width=0.8\linewidth]{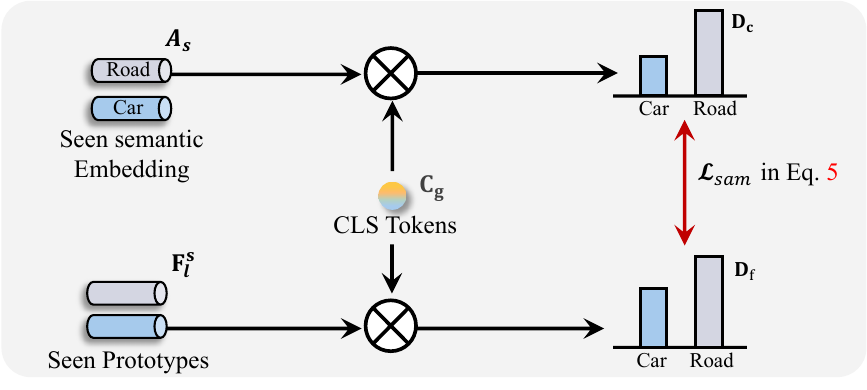}
\vspace{-0.1in}
\caption{The overview of \corr{Semantic Alignment Module (SAM)}.}
\label{fig:tam}
\vspace{-0.2in}
\end{figure}

\subsection{Selective Global Distillation (SGD)} \label{sec:SGD}
Though the proposed CSH enables the segmentation backbone to better align with CLIP's semantic space, relying solely on architectural design is insufficient to fully exploit the vision-language knowledge embedded in CLIP. To further facilitate the transfer of alignment capabilities, we introduce a \textbf{Selective Global Distillation (SGD)} module, as illustrated in Fig.~\ref{fig:SGD}. SGD aims to selectively align dense visual features with the CLS token from the CLIP visual encoder by focusing on regions that exhibit strong semantic similarity. This targeted distillation encourages the segmentation model to inherit CLIP's global semantic priors while avoiding noise from irrelevant or ambiguous regions.

Specifically, given the CLIP-bridged features $\textbf{F}_c \in \mathbb{R}^{H \times W \times D}$ from the CSH and a global CLS token $\textbf{C}_g \in \mathbb{R}^{D}$ extracted from the CLIP visual encoder, we compute similarity scores $\textbf{S} \in \mathbb{R}^{H \times W}$ between $\textbf{C}_g$ and each spatial location in $\textbf{F}_c$ using scaled dot-product similarity: $\textbf{S} = \frac{\textbf{F}_c^\top \textbf{C}_{g}}{\sqrt{D}}$. To prevent the model from repeatedly attending to the same highly activated regions, thus potentially overlooking semantically relevant areas with lower similarity scores, we adopt Gumbel-Softmax sampling to encourage diverse yet meaningful feature selection. Specifically, we perturb each similarity score $\textbf{S}$ with Gumbel noise and apply temperature-scaled softmax to obtain the sampling weights $\textbf{W}$:
\begin{equation}
g_i = -\log(-\log(\epsilon_i)), \quad \epsilon_i \sim \mathcal{U}(0,1),
\end{equation}
\begin{equation}
\mathbf{w}_i = \frac{\exp\left((s_i + g_i)/\tau\right)}{\sum_{j=1}^{HW} \exp\left((s_j + g_j)/\tau\right)} \quad s_i \in \textbf{S}, \, w_i \in \textbf{W},
\end{equation}
where $\mathcal{U}(0,1)$ denotes a uniform distribution. Based on the sampling weights $\textbf{W}$, we select the top-$K$ features to form a subset $\textbf{F}_k \in \mathbb{R}^{K \times D}$, and recompute similarity scores $\textbf{W}'$ between $\textbf{C}_g$ and $\textbf{F}_k$ using: $\textbf{W}' = \text{softmax}\left(\frac{\textbf{F}_k^\top \textbf{C}_g}{\sqrt{D}}\right)$. We then compute a global aggregated feature $\textbf{F}_g$ by weighting $\textbf{F}_k$ with $\textbf{W}_k$: $\textbf{F}_g = \textbf{W}_k^\top \textbf{F}_k$. InfoNCE is applied to align $\textbf{F}_g$ with $\textbf{C}_g$:
\begin{equation}
\mathcal{L}_{sgd} = \frac{\exp(\textbf{F}_g^{\top} \textbf{C}_g / \tau)}{\sum_{j = 1}^{B} \exp(\textbf{F}_g^{\top} \textbf{C}_j / \tau)},
\label{eq:tokendistillation}
\end{equation}
where $B$ denotes the batch size. As training progresses, we observe that the semantic representations of dense features become increasingly discriminative and diverse. To leverage this property, we introduce a \textit{\corr{decayed top-$K$ selecting}} strategy. Specifically, the number of selected features $K$ gradually decreases over time. In practice, we reduce $K$ by 0.1 per iteration and round it to the nearest integer. This decay mechanism encourages the model to focus on fewer but more semantically reliable regions as training proceeds. It also balances the trade-off between exploiting highly confident regions and exploring other potentially areas, thereby improving robustness against overfitting to dominant objects.

SGD not only mitigates the impact of noisy or ambiguous regions by directing distillation toward semantically salient features, but also enhances the model’s ability to inherit global vision-language alignment from CLIP. Through its \corr{decayed top-$K$ selecting} strategy, SGD progressively shifts attention toward the most informative regions, effectively balancing exploration and exploitation. This refinement improves the quality of feature alignment and promotes more robust generalization, ultimately leading to superior segmentation performance in challenging zero-shot settings.

\begin{figure}[t]
\centering
\includegraphics[width=0.9\linewidth]{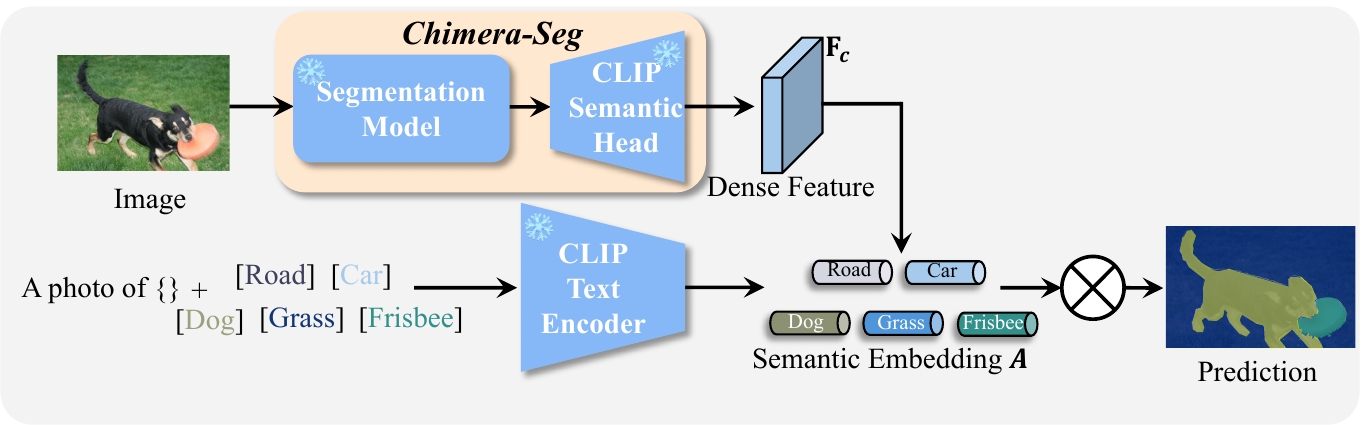}
\vspace{-0.05in}
\caption{The overview of inference.}
\label{fig:infer}
\vspace{-0.2in}
\end{figure}

\subsection{Training Objectives and Inference}
To obtain the final pixel-level logits, The final pixel-level logits is computed as $\textbf{P} = \textbf{F}_c^\top cat(\textbf{A}_s,\textbf{F}^u_l)$ where $cat$ indicates the concatenate along the class dimension. During training, the total loss can be represented as:
\begin{equation}
\begin{aligned}
  \mathcal{L} = &\mathcal{L}_{seg}(\textbf{P},\tilde{\textbf{Y}}) + \mathcal{L}_{sgd} + \lambda_{sam} \cdot \mathcal{L}_{sam}
\end{aligned}
\label{eq: total loss}
\end{equation}
where $\mathcal{L}_{seg}$ is the sum of focal loss \cite{focalloss} and cross entropy loss, respectively. $\mathcal{L}_{sam}$ indicates the loss of \corr{\textbf{Semantic Alignment Module (SAM)}} and the overview of $\mathcal{L}_{sam}$ is shown in Fig. \ref{fig:tam}. Given the seen semantic embeddings $\textbf{A}_s$ and the learned visual prototypes $\textbf{F}_l$, we first generate pseudo labels $\tilde{\textbf{Y}}$ via multi-scale K-Means and mask merging \cite{cliptoseg}, then compute per-class visual prototypes by averaging dense features across corresponding spatial regions,
\begin{equation}
\textbf{F}_l = \left\{f_l = \frac{\sum_{H,W}{\textbf{F}_c[\mathbb{1}(y_i = l)]}}{\sum_{H,W}{\mathbb{1}(y_i = l)}} \ \Big| \ y_i \in \tilde{\textbf{Y}} \right\},
\label{eq:visual prototype}
\end{equation}
where $\textbf{F}_l \in \mathbb{R}^{(O_s+O_u) \times D}$, $O_s$ and $O_u$ are the number of seen and latent classes in the input image and $\mathbb{1}(y_i = l)$ is an indicator function that selects pixels belonging to class $l$.
We split the prototypes $\textbf{F}_l$ into seen ($\textbf{F}_l^s$) and latent ($\textbf{F}_l^u$) groups. $\mathcal{L}_{sam}$ is defined as:
\begin{equation}
\mathcal{L}_{\text{sam}} = \text{KL}(\text{softmax}(\frac{\textbf{F}_l^s \cdot \textbf{C}_g^{\top}}{\tau_f}) \| \text{softmax}(\frac{\textbf{A}^s \cdot \textbf{C}_g^{\top}}{\tau_c}))),
\end{equation}
where $\tau_f$ and $\tau_c$ indicate two hyperparameters to control the sharpness of similarity distribution.

During inference, the overview is shown in Fig. \ref{fig:infer}. The input image is first encoded by the frozen segmentation model and projected into the CLIP-aligned semantic space via the CLIP-guided projector, resulting in dense visual features $\textbf{F}_c$. Simultaneously, the class names are converted into textual prompts (\eg, “A photo of [class]”) and embedded using the frozen CLIP text encoder to produce the semantic embedding matrix $\textbf{A}$. The final prediction is obtained by computing the similarity between each pixel-level feature in $\textbf{F}_c$ and all class embeddings in $\textbf{A}$, assigning each pixel to the most semantically similar class. Besides, under the inductive setting, we add a bias term $\gamma$ to the logits of unseen classes to calibrate their prediction confidence.

\begin{table*}[t]
\caption{Comparison under inductive settings where \textbf{bold} and \underline{underline} indicate the best and the second-best performance.}
\vspace{-5pt}
\setlength{\tabcolsep}{15pt}
\resizebox{\linewidth}{!}{
\begin{tabular}{cl|c|ccc|ccc}
\toprule
\multicolumn{2}{c|}{\multirow{2}{*}{Models}} & \multirow{2}{*}{Backbone}  & \multicolumn{3}{c|}{COCO-Stuff}               & \multicolumn{3}{c}{PASCAL Context}            \\ \cmidrule{4-9} 
\multicolumn{2}{c|}{}                        &                            & \textbf{sIoU} & \textbf{uIoU} & \textbf{hIoU} & \textbf{sIoU} & \textbf{uIoU} & \textbf{hIoU} \\ \midrule
\multicolumn{2}{c|}{SPNet \cite{spnet}}                   & \multirow{5}{*}{ResNet101 \cite{resnet}} & 35.2          & 8.7           & 14.0          & -             & -             & -             \\
\multicolumn{2}{c|}{ZS3 \cite{zs3}}                     &                            & 34.7          & 9.5           & 15.0          & 20.8          & 12.7          & 15.8          \\
\multicolumn{2}{c|}{CaGNet \cite{cagnet}}                  &                            & 33.5          & 12.2          & 18.2          & 24.1          & 18.5          & 21.2          \\
\multicolumn{2}{c|}{SIGN \cite{sign}}                    &                            & 32.3          & 15.5          & 20.9          & -             & -             & -             \\
\multicolumn{2}{c|}{Joint \cite{joint}}                   &                            & -             & -             & -             & 33.0          & 14.9          & 20.5          \\ \cmidrule{3-3}
\multicolumn{2}{c|}{ZegFormer \cite{zegformer}}               & \multirow{6}{*}{ViT-B \cite{vit}}     & 36.6          & 33.2          & 34.8          & -             & -             & -             \\
\multicolumn{2}{c|}{Zzseg \cite{simplebaseline}}                   &                            & 39.3          & 36.3          & 37.8          & -             & -             & -             \\
\multicolumn{2}{c|}{ZegCLIP \cite{zegclip}}                 &                            & 40.2          & 41.4          & 40.8          & 46.0          & 54.6          & 49.9          \\
\multicolumn{2}{c|}{DeOP \cite{DeOP}}                    &                            & 38.0          & 38.4          & 38.2          & -             & -             & -             \\
\multicolumn{2}{c|}{CLIP-RC \cite{clip-rc}}                 &                            & 40.9          & 41.6          & 41.2          & 47.5          & 57.3          & 51.9          \\
\multicolumn{2}{c|}{OTSeg+ \cite{otseg}}                  &                            & 41.3          & 41.8          & 41.5          & 55.2          & 60.4          & 57.7          \\ \cmidrule{3-3}
\multicolumn{2}{c|}{CLIP-to-Seg \cite{cliptoseg} }                & Segformer-B4 \cite{segformer}               & {\ul 43.2}    & {\ul 44.7}    & {\ul 43.9}    & {\ul 52.6}    & 64.5          & 58.0          \\ \midrule
\multicolumn{2}{c|}{\multirow{2}{*}{Ours}}   & ViT-B \cite{vit}                      & \textbf{43.4} & 44.1          & {\ul 43.7}    & 50.1          & 62.5          & 55.6          \\
\multicolumn{2}{c|}{}                        & Segformer-B4 \cite{segformer}             & 43.1          & \textbf{46.5} & \textbf{44.8} & \textbf{52.7} & \textbf{67.5} & \textbf{59.2} \\ \bottomrule
\end{tabular}
}
\label{tab: sota}
\end{table*}

\begin{table*}[t]
\caption{Comparison under transductive settings where \textbf{bold} and \underline{underline} indicate the best and the second-best performance where ``ST'' indicates the self-training.}
\vspace{-5pt}
\setlength{\tabcolsep}{15pt}
\resizebox{\linewidth}{!}{
\begin{tabular}{c|c|ccc|ccc}
\toprule
\multirow{2}{*}{Models} & \multirow{2}{*}{Backbone}  & \multicolumn{3}{c|}{COCO-Stuff}                                    & \multicolumn{3}{c}{PASCAL Context}            \\ \cmidrule{3-8} 
                        &                            & \textbf{sIoU} & \multicolumn{1}{c}{\textbf{uIoU}} & \textbf{hIoU} & \textbf{sIoU} & \textbf{uIoU} & \textbf{hIoU} \\ \midrule
SPNet+ST \cite{spnet}               & \multirow{6}{*}{ResNet101 \cite{resnet}} & 34.6          & 26.9                               & 30.3          & -             & -             & -             \\
ZS5  \cite{zs3}                   &                            & 34.9          & 10.6                               & 16.2          & 27.0          & 20.7          & 23.4          \\
CaGNet+ST  \cite{cagnet}             &                            & 35.6          & 13.4                               & 19.5          & -             & -             & -             \\
STRICT \cite{strict}                 &                            & 35.3          & 30.3                               & 34.8          & -             & -             & -             \\
FreeSeg  \cite{freeseg}               &                            & 42.2          & 49.1                               & 45.3          & -             & -             & -             \\
MaskCLIP+ \cite{maskclip}              &                            & 38.1          & 54.7                               & 45.0          & 44.4          & {\ul 66.7}    & 53.3          \\ \cmidrule{2-2}
Zzseg \cite{simplebaseline}                  & \multirow{4}{*}{ViT-B \cite{vit}}     & 39.6          & 43.6                               & 41.5          & -             & -             & -             \\
ZegCLIP+ST \cite{zegclip}              &                            & 40.7          & 59.9                               & 48.5          & 47.2          & 63.2          & 54.0          \\
OTSeg+ST \cite{otseg}               &                            & 41.4          & \textbf{62.6}                      & {\ul 49.8}    & \textbf{54.0} & {\ul 67.0}    & {\ul 59.8}    \\
CLIP-RC+ST \cite{clip-rc}             &                            & 42.0          & 60.8                               & 49.7          & 48.1          & 63.2          & 55.1          \\ \midrule
\multirow{2}{*}{Ours}   & ViT-B \cite{vit}                      & \textbf{44.4} & 59.7                               & \textbf{50.9} & 50.7          & 62.4          & 55.9          \\
                        & Segformer-B4 \cite{segformer}              & {\ul 43.7}    & {\ul 60.9}                         & \textbf{50.9} & {\ul 53.1}    & \textbf{72.5} & \textbf{61.3} \\ \bottomrule
\end{tabular}
}
\label{tab:sota trans}
\vspace{-0.2in}
\end{table*}

\section{Experiments}
\subsection{Dataset}
We validate the performance of \name~on two datasets: COCO-Stuff \cite{coco}, and PASCAL Context \cite{context}. Note that the split of seen and unseen classes follows the settings of the previous works \cite{zegclip,clip-rc}. Note that the `background' is converted to `ignored' in both training and inference. This dataset is split into 15 seen and 5 unseen classes. \textbf{COCO-Stuff} contains 118,287 training and 5,000 validation images. For ZSS, all 171 classes are grouped into 156 seen and 15 unseen classes. \textbf{PASCAL Context} maintains 4,996 training images and 5,104 testing images. In our experiment, the classes consist of 49 seen classes and 10 unseen classes. 

\subsection{Implementation Details}
Our method is based on MMSegmentation \cite{mmsegmentation} and trained on 8 V100 GPUs with 32GB memories per card. Unless specified, we use Segformer-B0 \cite{segformer} for most of the ablation studies. The visual encoder we use in CLIP is ViT-B/16 and the text encoder is the corresponding text encoder. For the hyperparameters, $\tau$ is set as 0.07, the $K$ in SGD is from 9000 and decreased 0.1 as each iteration, $\tau_c$ is set as 0.01, and $\tau_f$ is set as 0.07. The batch size we use is 16 and the model is trained in 40K and 80K iterations for PASCAL Context, and COCO-Stuff, respectively. $\lambda_{sam}$ are set as 0.1. Input images are resized to 512 $\times$ 512. Other hyperparameters follow the settings of CLIP2Seg \cite{cliptoseg} and the corresponding backbones. During inference in the COCO-Stuff dataset, we add a constant number $\gamma$ to the logits of unseen classes, and $\gamma$ is 0.5.

For ZSS, not only the seen classes but also the unseen classes should be considered. Therefore, following other researches \cite{cliptoseg,generalizableembedding,zegformer,zegclip}, we apply hIoU as the metric:
\begin{equation}
  hIoU = \frac{2 \cdot sIoU \cdot uIoU}{sIoU + uIoU},
  \label{eq:hiou}
\end{equation}
where $sIoU$ indicates the mIoU (mean Intersetcion-over-Union) for the seen classes and $uIoU$ indicates the mIoU for the unseen classes. We follow the zero-shot semantic segmentation (ZSS) setting, where annotations of unseen classes are removed from the training set while retaining all input images. We consider both inductive and transductive settings based on the availability of unseen class embeddings ($\mathbf{A}_u$) during training. In the inductive setting, $\mathbf{A}_u$ is not accessible, and unseen regions are treated as `ignored'. The transductive setting uses $\mathbf{A}_u$ for training. Evaluation covers both seen and unseen classes.

\begin{table}[t!]
\caption{Ablations on the proposed modules}
\vspace{-0.1in}
\setlength{\tabcolsep}{10pt}
\resizebox{\linewidth}{!}{
\begin{tabular}{lccc}
\toprule
\multicolumn{1}{c}{Module} & sIoU & uIoU & hIoU \\ \midrule
Baseline                   & 32.7 & 30.4 & 31.5 \\
Baseline + CSH             & 32.3 & 33.3 & 32.8 \\
Baseline + CSH + SGD       & 32.4 & 34.4 & 33.4 \\
Baseline + CSH + SGD + SAM & \textbf{32.5} & \textbf{35.2} & \textbf{33.8} \\ \bottomrule
\end{tabular}
}
\label{tab:ablations}
\vspace{-0.1in}
\end{table}

\begin{table}[t]
\caption{Ablations on the design of vision-text projector}
\vspace{-0.1in}
\setlength{\tabcolsep}{10pt}
\resizebox{\linewidth}{!}{
\begin{tabular}{ccc|cccc}
\toprule
Residual & BN  & VEncoder & pAcc & sIoU & uIoU & hIoU \\ \midrule
-        & -   & -        & 57.3 & 43.1 & 43.8 & 43.5 \\ 
-        & -   & \checkmark      & 31.9 & 15.1 & 17.9 & 16.4 \\
-        & \checkmark & -        & 62.1 & 43.0 & 44.4 & 42.7 \\
\checkmark      & \checkmark & -        & 61.7 & \textbf{43.5} & 44.4 & 44.0 \\
\checkmark      & -   & \checkmark      & 61.7    & 42.9    & 44.6    & 43.7    \\
-        & \checkmark & \checkmark      & \textbf{62.2} & 42.5 & 46.0 & 44.2 \\

\checkmark      & \checkmark & \checkmark      & 61.9 & 43.1 & \textbf{46.5} & \textbf{44.8} \\
\bottomrule
\end{tabular}
}
\label{tab: vt design}
\vspace{-0.2in}
\end{table}

\begin{table}[t]
\caption{Ablations on the projection layer}
\vspace{-5pt}
\setlength{\tabcolsep}{10pt}
\resizebox{\linewidth}{!}{
\begin{tabular}{cccccc}
\toprule
Initialization & Learnable & pAcc & sIoU & uIoU & hIoU \\ \midrule
CLIP           & -         & \textbf{56.1} & \textbf{32.5} & \textbf{35.2} & \textbf{33.8} \\
CLIP           & \checkmark       & 46.4 & 32.7 & 26.4 & 29.2 \\
He \cite{heinit}            & \checkmark       & 46.9 & 33.1 & 27.1 & 29.8 \\ \bottomrule
\end{tabular}}
\label{tab:proj}
\vspace{-0.2in}
\end{table}

\subsection{Comparison with State-of-the-Art}
Table~\ref{tab: sota} presents a comprehensive comparison of our method against state-of-the-art approaches on COCO-Stuff and PASCAL-Context under the Zero-Shot Semantic Segmentation (ZSS) setting. We report standard evaluation metrics: mIoU on seen classes (sIoU), unseen classes (uIoU), and their harmonic mean (hIoU). Our method consistently achieves the best performance across all metrics and benchmarks. On COCO-Stuff, our method achieves 43.4 sIoU / 44.1 uIoU / 43.7 hIoU with ViT-B and 43.1 sIoU / 46.5 uIoU / 44.8 hIoU with Segformer-B4, surpassing the previous state-of-the-art CLIP-to-Seg by a notable margin. On PASCAL-Context, our method further improves the state-of-the-art, achieving 52.7 sIoU / 67.5 uIoU / 59.2 hIoU with the Segformer-B4 backbone, outperforming CLIP-to-Seg (52.6 / 64.5 / 58.0). To ensure a fair comparison and isolate the contribution of our framework from that of the backbone architecture, we additionally conduct experiments using the ViT-B backbone, which is commonly adopted in prior CLIP-based segmentation works. Even under this setting, our model consistently outperforms existing methods such as ZegCLIP and OTSeg+, confirming that the observed improvements are attributed to our design rather than stronger vision backbones. These results highlight the effectiveness and generalizability of our proposed method, which enables efficient and robust cross-modal knowledge transfer by partially reusing CLIP’s visual and textual components in a flexible manner.

Table \ref{tab:sota trans} eports the performance of our method under the transductive setting on COCO-Stuff and PASCAL Context. Our approach achieves the best harmonic mean (hIoU) across both datasets, demonstrating strong generalization to unseen classes. Specifically, with Segformer-B4 as the backbone, our method attains 53.1 sIoU, 72.5 uIoU, and 61.3 hIoU on PASCAL Context, outperforming previous methods by a large margin. On COCO-Stuff, our model also achieves the best hIoU (50.9), validating its effectiveness in balancing seen and unseen class.

\subsection{Ablation Study} \label{sec: ablation}
We conduct the following ablations to validate our effectiveness. Unless specific, we use Segformer-B0 as the backbone for 80K iterations on COCO-Stuff.

\noindent \textbf{Ablations on the proposed modules.} We conduct a series of ablation studies to evaluate the effectiveness of each proposed component in our framework. As shown in Table~\ref{tab:ablations}, starting from a strong baseline, \ie, CLIP2Seg, we incrementally add the CLIP-Bridged Module (CSH), Selective Global Distillation (SGD), and Semantic Alignment Module (SAM). Introducing the CSH improves the model’s ability to generalize to unseen classes, as reflected by a notable increase in uIoU from 30.4 to 33.3. This demonstrates that leveraging partial CLIP layers can effectively bridge the semantic gap between vision and language. Adding the SGD module further boosts uIoU to 34.4 and hIoU to 33.4 by selectively distilling global knowledge from CLIP’s CLS token, thus refining the model’s semantic alignment. Finally, incorporating SAM yields the best overall performance, achieving 35.2 uIoU and 33.8 hIoU. This highlights the benefit of explicitly aligning visual prototypes with semantic embeddings under CLIP’s textual encoder.

\noindent \textbf{Design of vision-text projector and CLIP Projection.} We use Segformer-B4 as the backbone to validate each design of the vision-text projector. Table~\ref{tab: vt design} reports the ablation study on the components of the proposed CSH, including the residual connection, Batch Normalization (BN), and the VEncoder. Using only the frozen VEncoder severely degrades performance due to feature misalignment. Introducing BN improves stability, and the residual connection brings further gains. Combining both BN and residual yields a significant improvement (43.5 sIoU / 44.0 hIoU), confirming their complementary roles. The best performance (43.1 sIoU / 46.5 uIoU / 44.8 hIoU) is achieved when all components are used, validating the effectiveness of our CSH design for vision-language alignment.

\noindent \textbf{Ablations on the design of the CLIP projection.} In sec. \ref{sec:VT projector}, we assert that the frozen final CLIP projection is crucial for aligning vision and text. To evaluate the importance of the CLIP projection head, we conduct ablation studies under different initialization and training settings, as shown in Table~\ref{tab:proj}. Freezing the CLIP projection yields the highest performance (33.8 hIoU), highlighting its critical role in maintaining vision-language alignment. In contrast, making it learnable, whether initialized from CLIP or with Kaiming~\cite{heinit}, leads to a notable drop in uIoU and hIoU, suggesting that the learned parameters disrupt the semantic structure encoded in the original CLIP. These results support our claim that preserving the frozen projection is essential for effective zero-shot transfer.

\begin{figure*}[t!]
\centering
\includegraphics[width=1.0\linewidth]{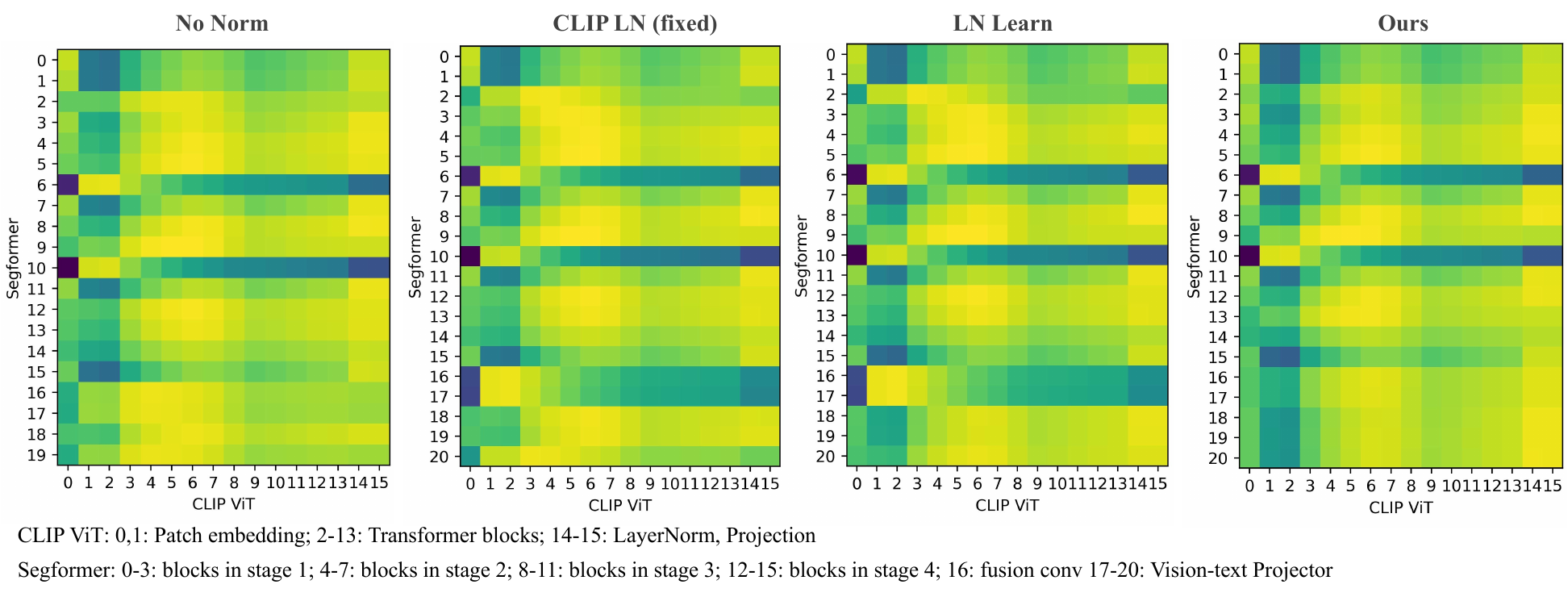}
\vspace{-0.2in}
\caption{Representation structure of ViT-CLIP vs. \name. We visualize four normalization types: no normalization, CLIP LN, learnable LN, and BN (ours). BN enables better integration of deeper semantic knowledge, while LN tends to preserve shallow, texture-level features.}
\vspace{-0.2in}
\label{fig: simi}
\end{figure*}

\begin{table}[t]
\caption{Ablations on the design of VEncoder}
\vspace{-5pt}
\setlength{\tabcolsep}{15pt}
\resizebox{\linewidth}{!}{
\begin{tabular}{cccccc}
\toprule
V   & FFN & pAcc & sIoU & uIoU & hIoU \\ \midrule
\checkmark & \checkmark & \textbf{56.1} & \textbf{32.5} & \textbf{35.2} & \textbf{33.8} \\
-   & \checkmark & 55.3 &\textbf{32.5} & 32.8 & 32.6 \\
\checkmark & -   & 55.7 & 32.5 & 34.0 & 33.2 \\
-   & -   & \textbf{56.0}    & 33.0    & 32.8    & 32.9    \\ \bottomrule
\end{tabular}}
\label{tab: vencoder design}
\vspace{-0.1in}
\end{table}

\begin{table}[t]
\caption{Ablations on learning of VEncoder, A-150 is ADE20K dataset}
\vspace{-5pt}
\setlength{\tabcolsep}{10pt}
\resizebox{\linewidth}{!}{
\begin{tabular}{ccccccc}
\toprule
V   & FFN & pAcc & sIoU & uIoU & hIoU & mIoU(A-150) \\ \midrule
-   & -   & \textbf{56.0} & 32.5 & \textbf{35.2} & 33.8 & \textbf{15.0}\\
-   & \checkmark & 55.8 & \textbf{33.3} & 35.0 & \textbf{34.1} & 14.8 \\
\checkmark & -   & 55.5 & 32.8 & 32.8 & 32.8 & 14.7 \\ \bottomrule
\end{tabular}}
\vspace{-0.2in}
\label{tab: vencoder learning}
\end{table}

\begin{table}[t]
\caption{Ablations on the number of VEncoder}
\vspace{-5pt}
\setlength{\tabcolsep}{15pt}
\resizebox{\linewidth}{!}{
\begin{tabular}{ccccc}
\toprule
VEncoder num& pAcc          & sIoU          & uIoU          & hIoU          \\ \midrule
0            & 56.0    & \textbf{33.0}    & 32.8    & 32.9             \\
1            & \textbf{56.1} & 32.5 & \textbf{35.2} & \textbf{33.8} \\
2            & 55.3          & 32.3          & 33.0          & 32.7          \\
3            & 55.6          & 32.2          & 33.1          & 32.6          \\ \bottomrule
\end{tabular}}
\label{tab: num vencoder}
\vspace{-0.1in}
\end{table}

\begin{table}[t]
\caption{Ablations on the choice of normalization}
\vspace{-5pt}
\setlength{\tabcolsep}{15pt}
\resizebox{\linewidth}{!}{
\begin{tabular}{ccccc}
\toprule
Norm      & pAcc          & sIoU          & uIoU          & hIoU          \\ \midrule
BN        & \textbf{56.1} & \textbf{32.5} & \textbf{35.2} & \textbf{33.8} \\
GN        & \textbf{56.0} & 32.3          & 34.6          & 33.4          \\
LN(CLIP)  & 53.6          & 31.4          & 29.3          & 30.3          \\
LN(learn) & 55.6          & 32.3          & 33.8          & 33.0          \\
None & 55.7          & 32.5          & 31.8          & 32.1          \\ \bottomrule
\end{tabular}}
\label{tab: norm}
\vspace{-0.2in}
\end{table}

\begin{table}[t]
\caption{Ablations on the way of decay in SGD}
\vspace{-5pt}
\setlength{\tabcolsep}{15pt}
\resizebox{\linewidth}{!}{
\begin{tabular}{ccccc}
\toprule
decay way & pAcc          & sIoU          & uIoU          & hIoU          \\ \midrule
no decay  & 55.7 & \textbf{32.4} & 34.2 & 33.3 \\
increase  & 55.9          & 32.3 & 34.3          & 33.3          \\
decrease  & \textbf{56.1}          & \textbf{32.5} & \textbf{35.2}          & \textbf{33.8}          \\ \bottomrule
\end{tabular}}
\label{tab: decay way}
\vspace{-0.1in}
\end{table}

\begin{table}[t]
\caption{Ablations on the number of decay number}
\vspace{-5pt}
\setlength{\tabcolsep}{15pt}
\resizebox{\linewidth}{!}{
\begin{tabular}{cccccc}
\toprule
num decay & pAcc          & sIoU          & uIoU          & hIoU & mIoU(ADE)          \\ \midrule
3000      & \textbf{56.1} & 32.4 & \textbf{35.1} & \textbf{33.7} & 14.9 \\
5000      & 56.0          & 32.4 & 34.7          & 33.5 & 14.9         \\
9000      & \textbf{56.1}          & \textbf{32.5} & \textbf{35.2}          & \textbf{33.8} & \textbf{15.0}          \\ \bottomrule
\end{tabular}}
\label{tab: decay num}
\vspace{-0.2in}
\end{table}

\begin{table}[t]
\caption{Ablations on $\lambda_{sam}$}
\vspace{-5pt}
\setlength{\tabcolsep}{20pt}
\resizebox{\linewidth}{!}{
\begin{tabular}{ccccc}
\toprule
$\lambda_{sam}$ & pAcc          & sIoU          & uIoU          & hIoU          \\ \midrule
0.0            & 55.9    & 32.4    & 34.8    & 33.6             \\
0.1            & 56.0 & 32.4 & 34.9 & 33.6 \\
0.5            & \textbf{56.1}          & \textbf{32.5}          & \textbf{35.2}          & \textbf{33.8}          \\
1.0            & 55.8          & 32.7          & 33.9          & 33.3          \\ \bottomrule
\end{tabular}}
\label{tab: ditillation loss}
\vspace{-0.2in}
\end{table}


\noindent \textbf{Ablations on the design of the VEncoder.} We investigate the contributions of different components in the proposed VEncoder. As shown in Table~\ref{tab: vencoder design}, removing the value projection (V) leads to a notable drop in hIoU (33.8\% → 32.6\%), confirming its essential role in vision-language alignment. Ablating the FFN results in a smaller drop, while completely removing VEncoder shows slightly better hIoU than using FFN alone, suggesting that the value path is the primary contributor, and FFN plays a complementary role.

We further examine the effect of freezing versus learning the VEncoder, as presented in Table~\ref{tab: vencoder learning}. Freezing both V and FFN achieves the best hIoU (33.8\%) and generalization on ADE20K (15.0\% mIoU). ADE20K \cite{ade20k} of 20K training images and 2K testing images with 150 classes as the benchmark. Note that all the scores related to this dataset are trained first on COCO-Stuff and tested on ADE20K without any finetuning. Making FFN learnable yields a marginal hIoU gain on COCO-Stuff but slightly degrades cross-dataset performance. When the value projection is also made learnable, performance drops on both benchmarks. These results highlight that freezing VEncoder provides strong semantic priors and better generalization across datasets.

\noindent \textbf{Ablations on the number of the VEncoder.} We further investigate how many VEncoder blocks should be used from the CLIP visual encoder. As shown in Table~\ref{tab: num vencoder}, using no VEncoder yields a strong baseline of 32.9\% hIoU. Adding one VEncoder block significantly improves uIoU (+2.4\%) and achieves the best overall performance (33.8\% hIoU). However, increasing the number of VEncoder blocks to two or three leads to performance degradation, suggesting that deeper layers introduce redundancy or harm alignment. These results confirm that the final block alone provides the most effective semantic bridge, consistent with findings in prior work~\cite{maskclip,sclip}.

\noindent \textbf{Ablations on the choice of normalization.} In Sec. \ref{sec:VT projector}, we argue that batch normalization is essential for the CSH. We ablate the normalization strategy used after the VEncoder in Table~\ref{tab: norm}. Batch Normalization (BN) yields the best overall performance with 35.2\% uIoU and 33.8\% hIoU, demonstrating strong stability and alignment with CLIP representations. Group Normalization (GN) performs slightly worse, while directly reusing CLIP’s LayerNorm results in a significant drop in uIoU and hIoU, indicating a distribution mismatch. Learnable LayerNorm achieves moderate results, and removing normalization entirely leads to degraded performance. These results highlight the effectiveness of BN in bridging backbone and CLIP feature distributions for zero-shot segmentation.

Furthermore, inspired by research analyzing representation \cite{dovision,cka}, we sought to understand why batch normalization (BN) performs better than layer normalization (LN). To investigate this, Fig. \ref{fig: simi}  shows the CKA similarity \cite{cka,dovision} between each layer of the SegFormer backbone and the CLIP ViT under four normalization settings: no normalization, fixed CLIP LayerNorm, learnable LayerNorm, and our proposed BatchNorm. A clear trend emerges: when LayerNorm (fixed or learnable) is used, the similarity between the final SegFormer layers (17–20, corresponding to the Vision-Text Projector) and the deeper CLIP layers (12–15) is significantly suppressed. This suggests that LayerNorm, originally designed for transformer training, may inhibit the alignment of deep semantic representations during cross-modal transfer. Interestingly, in the no normalization setting, we observe high similarity in deeper layers, but shallow layers become poorly aligned, indicating unstable or biased representation learning. In contrast, our BatchNorm design achieves a more balanced alignment across all stages, particularly enhancing similarity with the final CLIP layers (12–15), which are crucial for semantic alignment. These results confirm that BatchNorm not only stabilizes training but also facilitates the learning of CLIP’s high-level semantic structure, something LayerNorm tends to hinder in this cross-modal setup.

\begin{figure*}[t]
\centering
\includegraphics[width=0.9\linewidth]{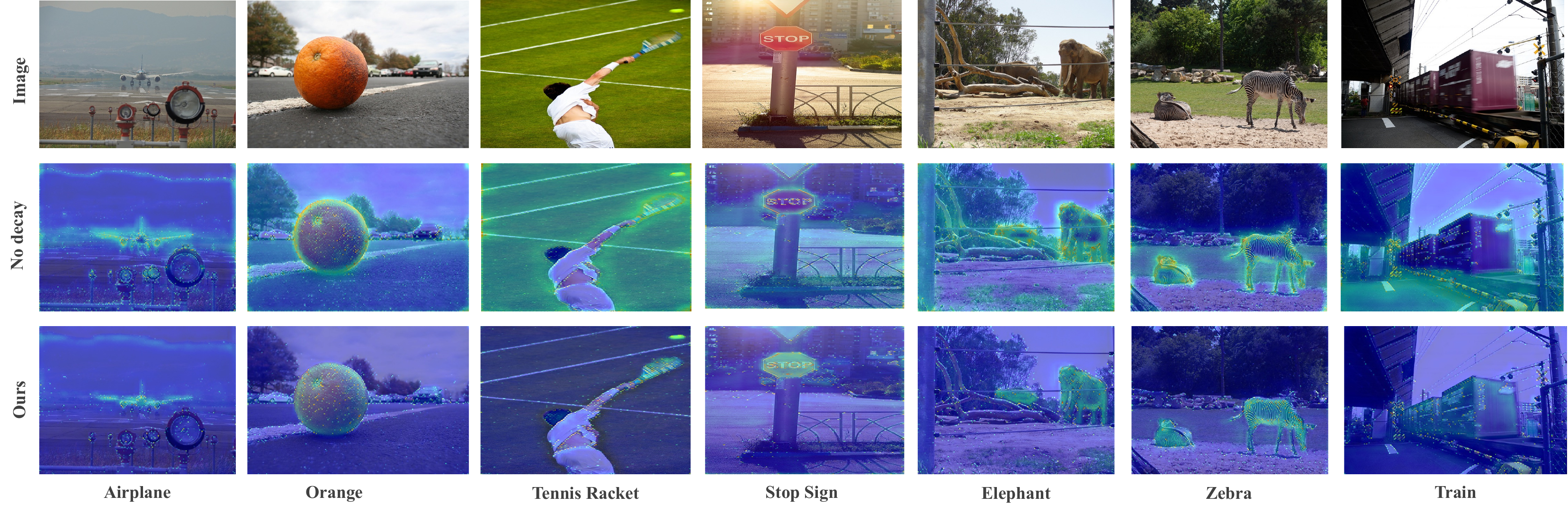}
\vspace{-0.1in}
\caption{Similarity visualization between CLS token and dense features. Each row corresponds to the input image, results without SGD, and with SGD, respectively. The labels beneath each image denote the semantic class of the corresponding CLS token.}
\label{fig:heat}
\vspace{-0.1in}
\end{figure*}


                     

                     

\begin{figure*}[t!]
\centering
\includegraphics[width=0.9\linewidth]{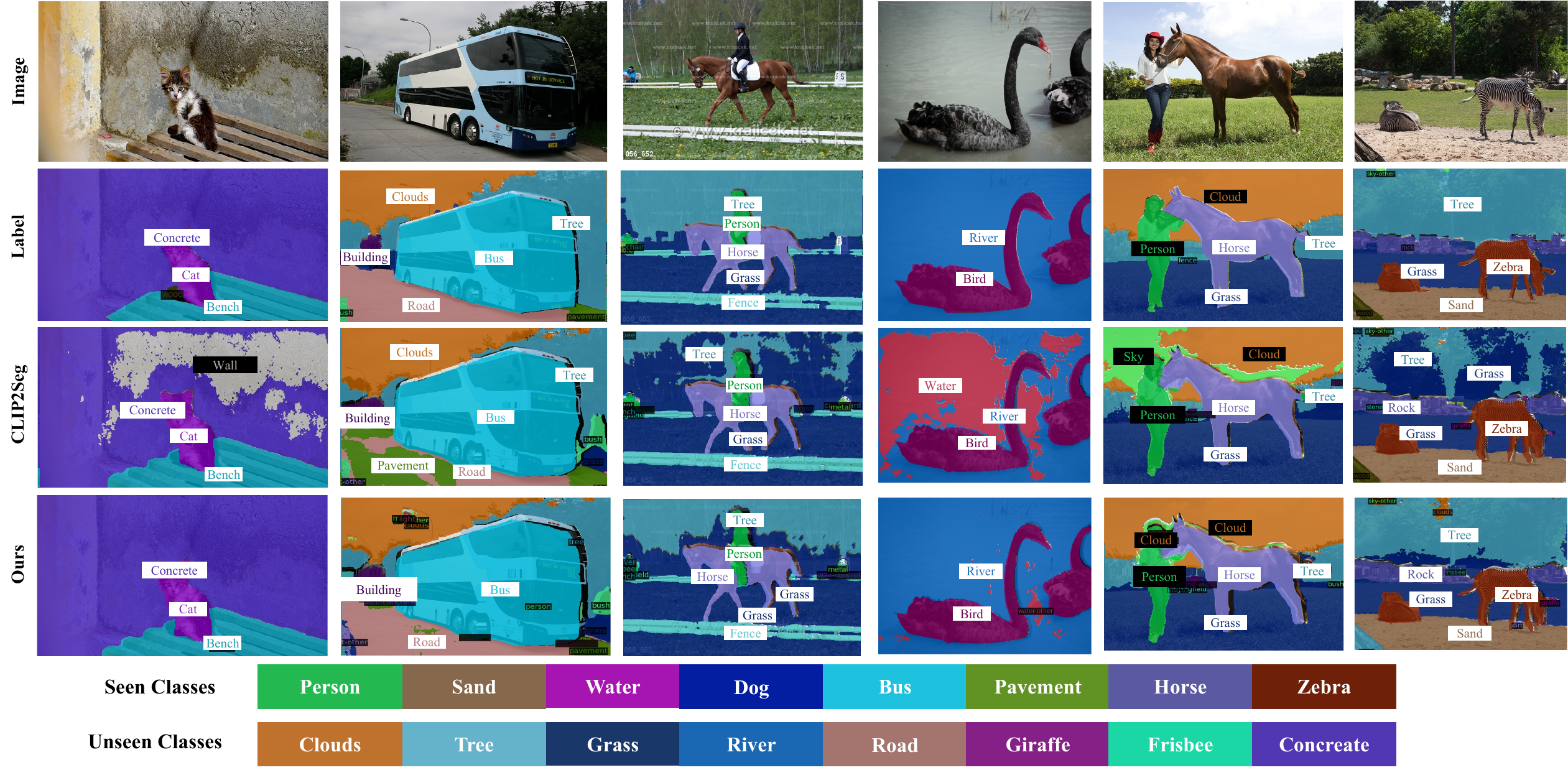}
\caption{The visualization of our method. We compare our method with the CLIP2Seg \cite{cliptoseg}.}
\label{fig:visualization}
\vspace{-0.2in}
\end{figure*}

\noindent \textbf{Ablations on the design of SGD.} In Sec. \ref{sec:SGD}, we declaim that the decayed $K$ is significant for solving the shortcut problem. We investigate the impact of different decay strategies in the Selective Global Distillation (SGD) module, as shown in Table~\ref{tab: decay way}. Without decay, the model achieves moderate performance, while increasing the number of selected features over training (increase) yields no improvement. In contrast, gradually decreasing the number of sampled features over time leads to the best results, achieving 35.2\% uIoU and 33.8\% hIoU. This suggests that selecting fewer but more reliable features as training progresses helps the model focus on semantically salient regions, improving alignment and generalization.

Besides, we visualize the effect of SGD through CLS-to-feature similarity maps. As shown in Fig.~\ref{fig:heat}, each map illustrates the similarity between a class-specific CLS token and dense features. Compared to the baseline without SGD, our method produces sharper and more semantically focused attention, highlighting target objects while suppressing irrelevant background regions. These results indicate that SGD encourages the CLS token to capture globally consistent, class-relevant cues, thereby improving vision-language alignment and segmentation quality.

Moreover, we aim to ablate the number of $K$ in the SGD as shown in Table~\ref{tab: decay num}. This table presents the effect of varying the total number of iterations over which top-$K$ sampling decays in the SGD module. We find that decaying from a higher value (\eg, 9000 pixels) consistently improves both uIoU and hIoU, achieving the best performance with 35.2\% uIoU and 33.8\% hIoU, along with the highest mIoU (15.0\%) on ADE20K. These results suggest that a gradual reduction in sampled regions allows the model to initially explore broader semantic cues and progressively focus on the most reliable regions, enhancing both alignment and generalization.

\begin{figure}[t!]
\centering
\includegraphics[width=0.9\linewidth]{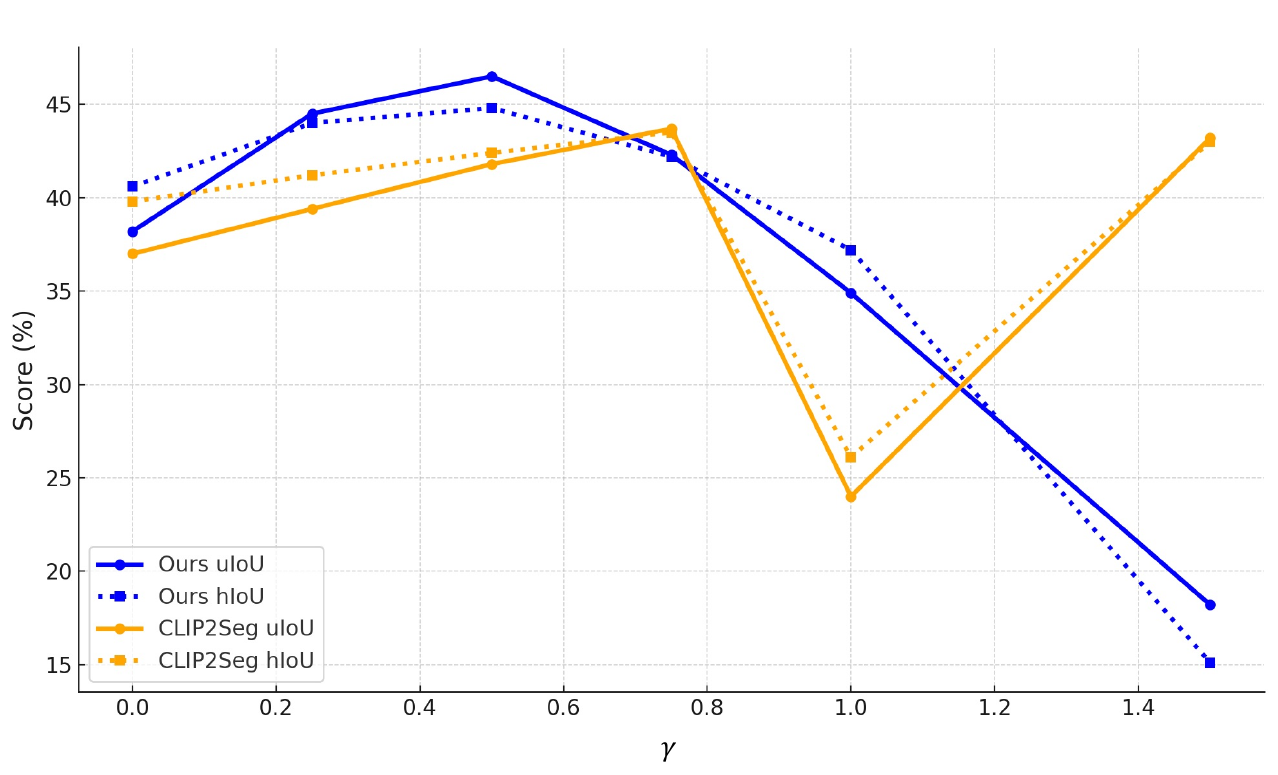}
\vspace{-0.1in}
\caption{Comparison on hIoU and uIoU between our method with CLIP2Seg.}
\label{fig:hiou}
\vspace{-0.2in}
\end{figure}

\noindent \textbf{Ablations on $\lambda_{sam}$.} Table~\ref{tab: ditillation loss} reports the impact of varying the loss weight $\lambda_d$ for the Selective Global Distillation (SGD) module. Setting $\lambda_d=0.5$ yields the best overall performance with 35.2\% uIoU and 33.8\% hIoU. Lower weights (\eg, 0.0 or 0.1) reduce the influence of the SGD objective, resulting in less effective feature alignment. On the other hand, overly large weights (\eg., 1.0) slightly degrade performance, likely due to the model overfitting to the global CLS token signal. These results suggest that a moderate distillation strength best balances segmentation and alignment objectives.


\noindent \textbf{Ablations on $\gamma$.} Fig.~\ref{fig:hiou} compares the impact of varying the hyperparameter $\gamma$ on unseen IoU (uIoU) and harmonic IoU (hIoU) between our method and CLIP2Seg. As $\gamma$ increases, our method initially shows consistent gains, peaking around $\gamma = 0.5$ with the highest uIoU and hIoU scores. However, performance drops sharply beyond this point, indicating sensitivity to excessive pseudo-label inclusion. In contrast, CLIP2Seg demonstrates greater robustness to $\gamma$ changes but achieves lower overall performance. These results highlight that while our approach benefits more from moderate pseudo-labels, it also requires careful calibration of $\gamma$ for stability.

\subsection{Qualitative Results}
To show the merits of our method, we visualize the prediction and compare them with the CLIP2Seg as shown in Fig. \ref{fig:visualization}. Compared with the baseline methods, \ie, CLIP2Seg, our method can better recognize the unseen classes, \eg, concrete in the first image and carrot in the third image, while maintaining the performance on the seen classes.

\section{Conclusion}
We propose \name, a plug-and-play framework that enables any segmentation model to acquire zero-shot capability by partially leveraging CLIP’s pretrained components. Unlike prior works that require full adaptation of vision-language models or rely solely on visual encoders, \corr{\name~bridges the gap between pure vision features and the CLIP-aligned space through the proposed CLIP Semantic Head (CSH) composed of partial modules from CLIP visual encoder and trainable components. The partial module from CLIP visual encoder, paired with the segmentation model, retains segmentation capability while easing the mapping to CLIP’s semantic space.} Furthermore, we introduce Selective Global Distillation (SGD) and the Semantic Alignment Module (SAM) to enhance vision-language alignment by selectively distilling discriminative features and explicitly incorporating textual semantics. Our method consistently improves zero-shot segmentation performance across multiple datasets, demonstrating both strong generalization and scalability. We believe \name~offers a practical and effective solution toward making segmentation models more annotation-efficient and semantically aware, and opens up new directions for integrating vision-language knowledge into pixel-level tasks.

\noindent \textbf{Limitation and Future Works:} While effective, our method still has some limitations. Although CSH enhances the model’s ability to capture semantics from deeper layers of the CLIP visual encoder, the underlying mechanism behind this improvement remains underexplored. Furthermore, although SGD leads to more semantically aligned similarity maps between dense features and the CLIP CLS token, it occasionally activates irrelevant regions. 

\section*{Acknowledgment}
Support for this work was given by the Toyota Motor Corporation (TMC) and JSPS KAKENHI Grant Number 23K28164 and JST CREST Grant Number JPMJCR22D1. However, note that this article solely reflects the opinions and conclusions of its authors and not TMC or any other Toyota entity. Computations are done on supercomputer “Flow” at the Information Technology Center, Nagoya University.

\bibliographystyle{IEEEtran}
\bibliography{ref}

\end{document}